\documentclass[letterpaper, 10 pt, conference]{ieeeconf}  

\IEEEoverridecommandlockouts                              

\usepackage{subcaption}
\usepackage{graphicx}
\usepackage{algorithm,algorithmicx, algpseudocode}
\usepackage{tikz}
\usepackage{verbatim}
\usepackage{amsmath, amssymb}
\usepackage{pgfplots}
\usetikzlibrary{arrows.meta}
\usepackage{url}

\title{\LARGE \bf Learning to Drive in a Day}

\author{Alex Kendall \quad Jeffrey Hawke \quad David Janz \quad Przemyslaw Mazur \quad Daniele Reda\\John-Mark Allen \quad Vinh-Dieu Lam \quad Alex Bewley \quad Amar Shah%
\thanks{The authors are with Wayve in Cambridge, UK.}
\thanks{{\tt\small research@wayve.ai}}%
}

\graphicspath{{./figures/}}

\newcommand{\argmax}{\mathrm{argmax}}

\begin{document}

\maketitle
\thispagestyle{empty}
\pagestyle{empty}

\begin{abstract}

We demonstrate the first application of deep reinforcement learning to autonomous driving. From randomly initialised parameters, our model is able to learn a policy for lane following in a handful of training episodes using a single monocular image as input. We provide a general and easy to obtain reward: the distance travelled by the vehicle without the safety driver taking control. We use a continuous, model-free deep reinforcement learning algorithm, with all exploration and optimisation performed on-vehicle. This demonstrates a new framework for autonomous driving which moves away from reliance on defined logical rules, mapping, and direct supervision. We discuss the challenges and opportunities to scale this approach to a broader range of autonomous driving tasks.

\end{abstract}

\section{Introduction}

Autonomous driving is a topic that has gathered a great deal of attention from both the research community and companies, due to its potential to radically change mobility and transport. Broadly, most approaches to date focus on formal logic which define driving behaviour in annotated 3D geometric maps. This can be difficult to scale, as it relies heavily on external mapping infrastructure rather than primarily using an understanding of the local scene.


In order to make autonomous driving a truly ubiquitous technology, we advocate for robotic systems which address the ability to drive and navigate in absence of maps and explicit rules, relying - just like humans - on a comprehensive understanding of the immediate environment~\cite{badrinarayanan2017segnet} while following simple higher level directions (e.g., turn-by-turn route commands). Recent work in this area has demonstrated that this is possible on rural country roads, using GPS for coarse localisation and LIDAR to understand the local scene~\cite{ort2018autonomous}.

In the recent years, reinforcement learning (RL) -- a machine learning subfield focused on solving Markov Decision Problems (MDP)~\cite{sutton1998rli} where an agent learns to select actions in an environment in an attempt to maximise some reward function -- has shown an ability to achieve super-human results at games such as Go~\cite{silver2016go} or chess~\cite{silver2017chess}, a great deal of potential in simulated environments like computer games~\cite{mnih2015human}, and on simple tasks with robotic manipulators~\cite{gu2016robotic}. We argue that the generality of reinforcement learning makes it a useful framework to apply to autonomous driving. Most importantly, it provides a corrective mechanism to improve learned autonomous driving behaviour.

\begin{figure}
\centering
\includegraphics[width=\linewidth]{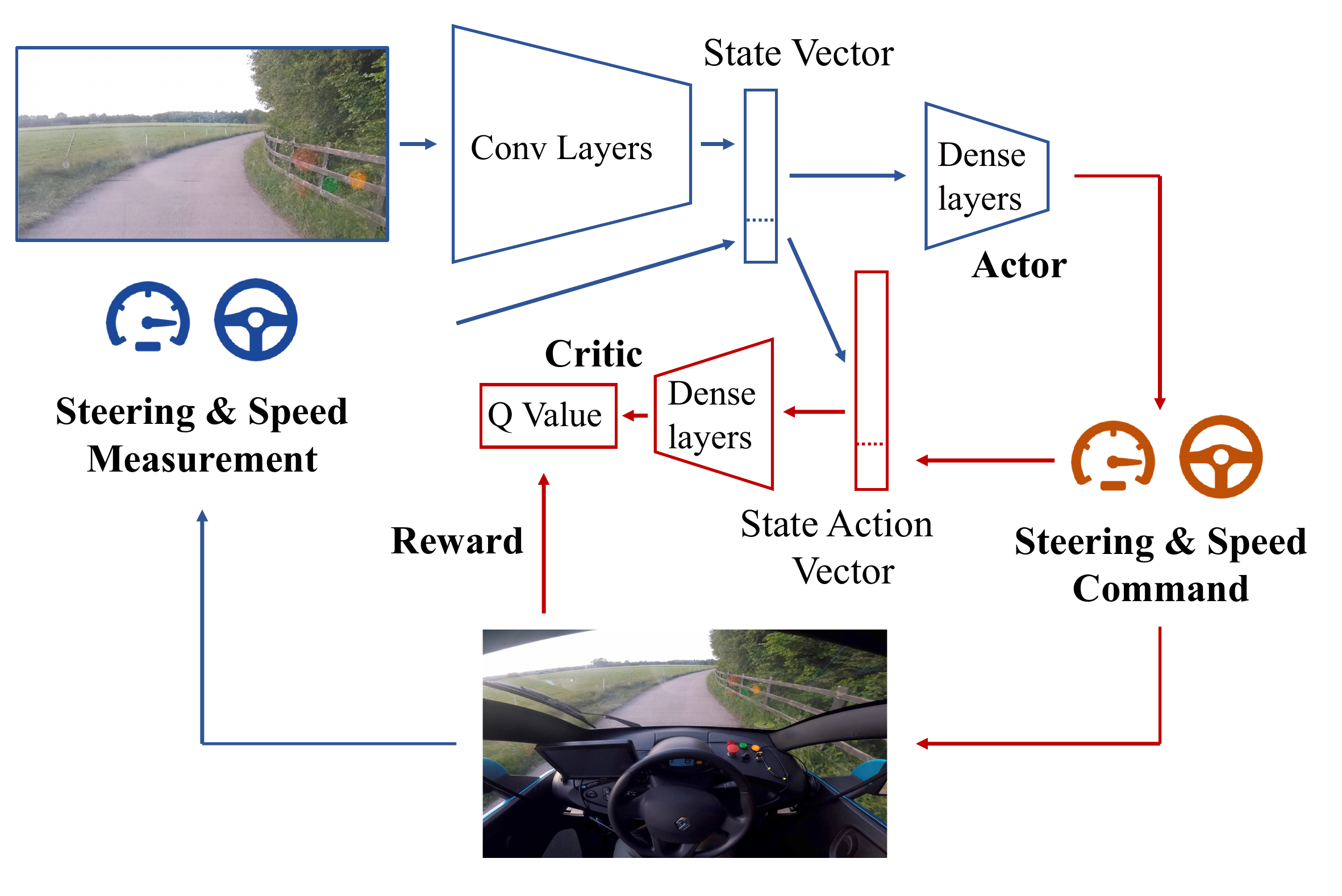}
\caption{We design a deep reinforcement learning algorithm for autonomous driving. This figure illustrates the actor-critic algorithm which we use to learn a policy and value function for driving. Our agent maximises the reward of distance travelled before intervention by a safety driver. A video of our vehicle learning to drive is available at \protect\url{https://wayve.ai/blog/l2diad}}
\label{fig:teaser}
\end{figure}

To this end, in this paper we:
\begin{enumerate}
\item pose autonomous driving as an MDP, explain how to design the various elements of this problem to make it simpler to solve, whilst keeping it general and extensible,
\item show that a canonical RL algorithm (deep deterministic policy gradients \cite{lillicrap2015continuous}) can rapidly learn a simple autonomous driving task in a simulation environment,
\item discuss the system set-up required to make learning to drive efficient and safe on a real-world vehicle,
\item learn to drive a real-world autonomous vehicle in a few episodes with a continuous deep reinforcement learning algorithm, using only on-board computation.
\end{enumerate}
We therefore present the first demonstration of a deep reinforcement learning agent driving a real car.


\section{Related Work}

We believe this is the first work to show that deep reinforcement learning is a viable approach to autonomous driving. We are motivated by its potential to scale beyond that of imitation learning, and hope the research community examines autonomous driving from a reinforcement learning perspective more closely. The closest work in the current literature can predominantly be categorised as either imitation learning or classical approaches relying on mapping.

\paragraph{Mapping approaches}
Since early examples~\cite{kanade1986autonomous,wallace1985first}, autonomous vehicle systems have been designed to navigate safely through complex environments using advanced sensing and control algorithms~\cite{montemerlo2008junior,levinson2011towards,franke1998autonomous}. These systems are traditionally composed of many specific independently engineered components, such as perception, state estimation, mapping, planning and control~\cite{thrun2005probabilistic}. However, because each component needs to be individually specified and tuned, this can be difficult to scale to more difficult driving scenarios due to complex interdependencies.

Significant effort has been focused on computer vision components for this modular approach. Localisation such as~\cite{LinegarICRA2016} facilitates control of the vehicle~\cite{muller2006off} within the mapped environment, while perception methods such as semantic segmentation~\cite{badrinarayanan2017segnet} enable the robot to interpret the scene. These modular tasks are supported by benchmarks such as~\cite{geiger2012we} and~\cite{MaddernIJRR2016}.

These modular mapping approaches are largely the focus of commercial efforts to develop autonomous driving systems; however, they present an incredibly complex systems engineering challenge, which has yet to be solved.

\paragraph{Imitation learning}
A more recent approach to some driving tasks is imitation learning~\cite{pomerleau1989alvinn,bojarski2016end}, which aims to learn a control policy by observing expert demonstrations. One important advantage of this approach is that it can use end-to-end deep learning, optimising all parameters of a model jointly with respect to an end goal thus reducing the effort of tuning of each component. However, imitation learning is also challenging to scale. It is impossible to obtain expert examples to imitate for every potential scenario an agent may encounter, and it is challenging to deal with distributions of demonstrated policies (e.g., driving in each lane). 

\paragraph{Reinforcement learning}
Reinforcement learning is a broad class of algorithms for solving Markov Decision Problems (MDPs)~\cite{sutton1998reinforcement}. An MDP consists of:
\begin{itemize}
 \item a set $\mathcal{S}$ of states,
 \item a set $\mathcal{A}$ of actions,
 \item a transition probability function $p \colon \mathcal{S} \times \mathcal{A} \to \mathcal{P}(\mathcal{S})$, which to every pair $(s, a) \in \mathcal{S} \times \mathcal{A}$ assigns a probability distribution $p(\cdot | s, a)$ representing the probability of entering a state from state $s$ using action $a$,
 \item a reward function $R \colon \mathcal{S} \times \mathcal{S} \times \mathcal{A} \to \mathbb{R}$, which describes the reward $R(s_{t+1}, s_t, a_t)$ associated with entering state $s_{t+1}$ from state $s_t$ using action $a_t$,
 \item a future discount factor $\gamma \in [0, 1]$ representing how much we care about future rewards.
\end{itemize}

The solution of an MDP is a policy $\pi\colon \mathcal{S} \to \mathcal{A}$ that for every $s_0 \in \mathcal{S}$ maximises:

\begin{equation}
 V_{\pi}(s_0) = \mathbb{E} \left( \sum_{t=0}^\infty \gamma^t R(s_{t+1}, s_t, \pi(s_t)) \right),
\end{equation}
where the expectation is taken over states $s_{t+1}$ sampled according to $p(s_{t+1} | s_t, \pi(s_t))$.

In our setting, we use a finite time horizon $T$ in place of infinity in the above formula. This is equivalent to one of the states being terminal, i.e. it cannot be escaped and any action at that state gives zero reward.

Rearranging the above equation into a recurrent form, we get one of the two Bellman equations:

\begin{equation}
 V_{\pi}(s_0) = \mathbb{E} \bigg( R(s_1, s_0, \pi(s_0)) + \gamma V_\pi(s_1) \bigg).
\end{equation}

Here the expectation is taken only over $s_1$ sampled according to $p(s_1 | s_0, \pi(s_0))$. For reference, let us present the other Bellman equation:

\begin{equation}
 Q_{\pi}(s_0, a_0) = \mathbb{E} \bigg( R(s_1, s_0, a_0) + \gamma Q_\pi(s_1, \pi(s_1)) \bigg),
\end{equation}
where $Q_{\pi}(s_0, a_0)$ is the expected cumulative discounted reward received while starting from state $s_0$ with action $a_0$ and following policy $\pi$ thereafter. Again  the expectation is taken over $s_1$ sampled according to $p(s_1 | s_0, a_0)$

In other words, reinforcement learning algorithms aim to learn a policy $\pi$ that obtains a high cumulative reward. They are generally split into two categories: model-based and model-free reinforcement learning. In the former approach, explicit models for the transition and reward functions are learnt, and then used to find a policy that maximises cumulative reward under those estimated functions. In the latter, we directly estimate the value $Q(s, a)$ of taking action $a$ in state $s$, and then follow a policy that selects the action with the highest estimated value in each state.

Model-free reinforcement learning is extremely general. Using it, we can (in theory) learn any task we can imagine, whereas model-based algorithms can be only as good as the model learned. On the other hand, model-based methods tend to be more data-efficient than model-free ones. For further discussion, see~\cite{deisenroth2011pilco}.

In autonomous driving, deep learning has been used to learn dynamics models for model-based reinforcement learning using off-line data \cite{williams2017information}. Reinforcement learning has also been used to learn autonomous driving agents in video games. However, this can simply the problem, with access to ground truth reward signals which are not available in the real-world, such as the angle of the car to the the lane \cite{lillicrap2015continuous}.	

The closest work to this paper is from Riedmiller et al.~\cite{riedmiller2007learning} who train a reinforcement learning agent which drives a vehicle to follow a GPS trajectory in an obstacle-free environment. They demonstrate learning on-board the vehicle using a dense reward function based on GPS thresholded tracking error. We build on this work in a number of ways; we demonstrate learning to drive with deep learning, from an image-based input, using a sparse reward function to lane follow.


\section{System Architecture}

\subsection{Driving as a Markov Decision Process}
\label{section:mdp}
A key focus of this paper is the set-up of driving as an MDP. Our goal is that of autonomous driving, and the exact definition of the state space $\mathcal{S}$, action space $\mathcal{A}$ and reward function $R$ are free for us to be defined. The transition model is implicitly fixed once a state and action representation is fixed, with the remaining degrees of freedom -- the transitions themselves -- dictated by the mechanics of the simulator/vehicle used.

\paragraph{State space} Key to defining the state space is the definition of the observations $O_t$ that the algorithm receives at each time step. Many sensors have been developed in order to provide sophisticated observations for driving algorithms, not limited to LIDAR, IMUs, GPS units and IR depth sensors; an endless budget could be spent on advanced sensing technology. In this paper, we show that for simple driving tasks it is sufficient to use a monocular camera image, together with the observed vehicle speed and steering angle. Theoretically, state $s_t$ is to be a Markov representation of all previous observations. An approximation a fixed length approximately Markov state could be obtained by, for example, using a Recurrent Neural Network to recursively combine observations. However, for the tasks we consider, the observation itself serves as a good enough approximation of the state.

A second consideration is how to treat the image itself: the raw image could be fed directly into the reinforcement learning algorithm through a series of convolutions~\cite{lecun1989backpropagation}; alternatively, a small compressed representation of the image, using, for example, a Variational Autoencoder (VAE)~\cite{kingma2013auto}~\cite{rezende2014stochastic}, could be used. We compare the performance of reinforcement learning using these two approaches in Section~\ref{section:experiments}. In our experiments, we train the VAE online from five purely random exploration episodes, using a KL loss and a L2 reconstruction loss~\cite{rezende2014stochastic}.

\paragraph{Action space} Driving itself has what one might think are a natural set of actions:
throttle, brake, signals etc. But what domain should the output of the reinforcement learning algorithm be? The throttle itself can be described as discrete, either on or off, or continuous, in a range isometric to [0, 1]. An alternative is to reparameterise the throttle in terms of a speed set-point, with throttle output by a classical controller in an attempt to match the set-point. Overall, experiments on a simple simulator (Section~\ref{section:sim}) showed that continuous actions, whilst somewhat harder to learn, provide for a smoother controller. We use a two-dimensional action space; steering angle in the range [-1, 1] and speed setpoint in km/h.

\paragraph{Reward function} Design of reward functions can approach supervised learning -- given a lane classification system, a reward to learn lane-following can be set up in terms of minimising the predicted distance from centre of lane, the approach taken in~\cite{lillicrap2015continuous}. This approach is limited in scale: the system can only be as good as the human intuition behind the hand-crafted reward. We do not take this approach. Instead, we define the reward as forward speed and terminate an episode upon an infraction of traffic rules -- thus the value of a given state $V(s_t)$ corresponds to the average distance travelled before an infraction. A fault that may be identified is that the agent may choose to avoid more difficult manoeuvres, e.g. turning right in the UK (left in US). Command conditional rewards may be utilised in future work to avoid this.

\subsection{Reinforcement Learning Algorithm -- Deep Deterministic Policy Gradients}
We selected a simple continuous action domain model-free reinforcement learning algorithm: deep deterministic policy gradients (DDPG)~\cite{lillicrap2015continuous}, to show that an off-the-shelf reinforcement learning algorithm with no task-specific adaptation is capable of solving the MDP posed in Section~\ref{section:mdp}.

DDPG consists of two function approximators: a critic $Q \colon \mathcal{S} \times \mathcal{A} \to \mathbb{R}$, which estimates the value $Q(s, a)$ of the expected cumulative discounted reward upon using action $a$ in state $s$, trained to satisfy the Bellman equation
\begin{equation*}
 Q(s_t, a_t) = r_{t+1} + \gamma (1 - d_t) Q(s_{t+1}, \pi(s_{t+1})),
\end{equation*}
under a policy given by the actor $\pi \colon \mathcal{S} \to \mathcal{A}$, which attempts to estimate a $Q$-optimal policy $\pi(s) = \argmax_a Q(s, a)$; here $(s_t, a_t, r_{t+1}, d_{t+1}, s_{t+1})$ is an experience tuple, a transition from state $s_t$ to $s_{t+1}$ using action $a_t$ and receiving reward $r_{t+1}$ and ``done'' flag $d_{t+1}$, selected from a buffer of past experiences. The error in the Bellman equality, which the critic attempts to minimise, is termed the temporal difference ($TD$) error. Many variants of actor-critic methods exist, see e.g.~\cite{sutton2000policy, mnih2016asynchronous}.

DDPG training is done online. Beyond the infrastructure of setting up such a buffer for use on a real vehicle (which requires it to be tolerant of missing/faulty episodes and any-time stoppable), reinforcement learning can be sped up by selecting the most ``informative'' examples from the replay buffer. We do so using a commonly established method called prioritised experience replay~\cite{schaul2015prioritized}: we sample experience tuples with probability proportional to the $TD$ error made by the critic. The weights used for this sampling are updated upon each optimisation step with minimal overhead; new samples are given infinite weight to ensure all samples are seen at least once.

DDPG is an off-policy learning algorithm, meaning that actions performed during training come from a policy distinct from the learn optimal policy by the actor. This happens in order to gain diverse state-action data outside of the narrow distribution that would be seen by the optimal policy, and thus increase robustness. We use a standard method of achieving this in the context of continuous reinforcement learning methods: our exploration policy is formed by adding discrete Ornstein-Uhlenbeck process noise~\cite{uhlenbeck1930brownian} to the optimal policy. Therefore, at each step we add to optimal actions noise $x_t$ given by:
\begin{equation}
  \label{eq:1}
  x_{t+1} = x_t + \theta (\mu - x_t) + \sigma \epsilon_t,
\end{equation}
where $\theta, \mu, \sigma$ are hyperparameters and $\{\epsilon_t\}_t$ are i.i.d. random variables sampled from the normal distribution $N(0, 1)$. These parameters need to be tuned carefully, as there is a direct trade-off between noise utility and comfort of the safety driver. Strongly mean reverting noise with lower variance is easier to anticipate, whilst higher variance noise provides better state-action space coverage.

\begin{figure*}
\centering
	\begin{subfigure}[b]{0.49\textwidth}
		\begin{algorithmic}[1]
		\While{True}
		\State Request \texttt{task}
		\State Waiting for environment reset
		\If{\texttt{task} is train}
		\State Run episode with \textit{noisy policy}
		\If{exploration time is over}
		\State Optimise model
		\EndIf
		\ElsIf{\texttt{task} is test}
		\State Run episode with \textit{optimal policy}
		\ElsIf{\texttt{task} is undo}
		\State Revert previous train/test task
		\ElsIf{\texttt{task} is done}
		\State Exit experiment
		\EndIf
		\EndWhile
		\end{algorithmic}
		\caption{Task-based workflow for on-vehicle training}
		\label{alg:rl-training-logic}
     \end{subfigure}
     \begin{subfigure}[b]{0.5\textwidth}
		\centering
		\resizebox{5cm}{!}{%
		    \tikzstyle{block} = [rectangle, draw, text width=8em, text centered,
		    rounded corners, minimum height=3em]
		    \begin{tikzpicture}[align = center, node distance = 8em, auto, thick]
		        \node [block] (Trainer) {Stateful Trainer};
		        \node [block, below of=Trainer, yshift=+0.5cm] (Model) {RL Model};
		        \node [block, below of=Model, yshift=+0.5cm] (Controller) {Controller};
		        \node [block, below of=Controller, yshift=+0.5cm] (Vehicle) {Vehicle};

		        \node [left of=Model, xshift=0.5cm] (t1) {\includegraphics[width=.05\textwidth]{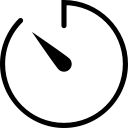}\\10Hz};
		        \node [left of=Controller, xshift=0.5cm] (t2)  {\includegraphics[width=.05\textwidth]{clock.png}\\100Hz};

		        \coordinate [right of=Model, xshift=0.5cm] (t1);
		        \coordinate [right of=Controller, xshift=0.5cm] (t2);
		        \coordinate [right of=Vehicle, xshift=0.5cm] (t3);

		        \draw [-{Latex[width=2mm]}] (Trainer) edge node[right] {Task} (Model);
		        \draw [-{Latex[width=2mm]}] (Model) edge node[right] {Action} (Controller);
		        \draw [-{Latex[width=2mm]}] (Controller) edge node[right] {Control Actuation} (Vehicle);
		        \draw [-{Latex[width=2mm]}] (t1) -- node[above] {State} node[below] {(w. Images)} (Model);
		        \draw [-{Latex[width=2mm]}] (t2) -- node[above] {State} node[below] {(Telemetry)} (Controller);
		        \draw [-] (t3) -- (Vehicle);
		        \draw [-] (t1) -- (t3);
		    \end{tikzpicture}
		}
		\caption{Policy execution architecture, used to run episodes during model training or testing.}
		\label{fig:rl-training-logic}
    \end{subfigure}
\caption{Outline of the workflow and the architecture for efficiently training the algorithm from a safety driver's feedback.}
\end{figure*}

\subsection{Task-based Training Architecture}
\label{section:stateful}
Deployment of a reinforcement learning algorithm on a full-sized robotic vehicle running in a real world environment requires adjustment of common training procedures, to account for both driver intervention and external variables affecting the training.

We structure the architecture of the algorithm as a simple state machine, outlined in Figure~\ref{alg:rl-training-logic}, in which the safety driver is in control of the different tasks. We define four tasks: train, test, undo and done. The definition of these tasks allows the system to be both interactive and stateful, favouring an on-demand execution of episodes instead of an a priori fixed schedule.

The train and test tasks allow us to interact with the vehicle in autonomous mode, executing the current policy. The difference between the two tasks consists in noise being added to the model output and the model being optimised in training tasks, whereas test tasks run directly the model output actions. During early episodes, we skip optimisation to favour exploration of the state space. We continue the experiment until the test reward stops increasing.

Each episode is executed until the system detects that automation is lost (i.e. the driver intervened). In a real world environment, the system can not reset automatically between episodes, unlike agents in simulation or in a constrained environment. We require a human driver to reset the vehicle to a valid starting state. Upon episode termination, while the safety driver performs this reset, the model is being optimised, minimising the time spent between episodes.

The undo and done tasks depict the key differences in the architecture. The system may terminate an episode for a variety of valid reasons other than failing to drive correctly: these episodes can not be considered for the purposes of training. The undo task is introduced for this reason, as it allows us to undo the episode and restore the model as it was before running that episode. A common example in our experiments is encountering other drivers seeking to use the road being used as the environment. The done task allows us to gracefully exit the experiment at any given moment, and is helpful since the procedure is interactive and it doesn't run for a fixed number of episodes.


\section{Experiments}
\label{section:experiments}
The main task we use to showcase the vehicle is that of lane-following; this is the same task as addressed in~\cite{lillicrap2015continuous}, however done on a real vehicle as well as on simulation, and done from image input, without knowledge of lane position. It is a task core to driving, and was the cornerstone of the seminal ALVINN~\cite{pomerleau1989alvinn}. We first accomplish this task in simulation in Section~\ref{section:sim}, and then use these results and knowledge of appropriate hyperparameters to demonstrate a solution on a real vehicle in Section~\ref{section:real_world}.

For both simulation and real-world experiments we use a small convolutional neural network. Our model has four convolutional layers, with $3 \times 3$ kernels, stride of 2 and 16 feature dimensions, shared between the actor and critic models. We then flatten the encoded state and concatenate the vector the scalar state for the actor, additionally concatenating the actions for the critic network. For both networks we then apply one fully-connected layer with feature size 8 before regressing to the output. For the VAE experiments, a decoder of the same size as the encoder is used, replacing strided convolution with transposed convolution to upsample the features. A graphical depiction is shown in Figure~\ref{fig:teaser}.

\subsection{Simulation}
\label{section:sim}
\begin{figure}[t]
\centering
     \begin{subfigure}[b]{0.4\linewidth}
         \centering
         \includegraphics[width=\linewidth,trim={0 0 0 50px},clip]{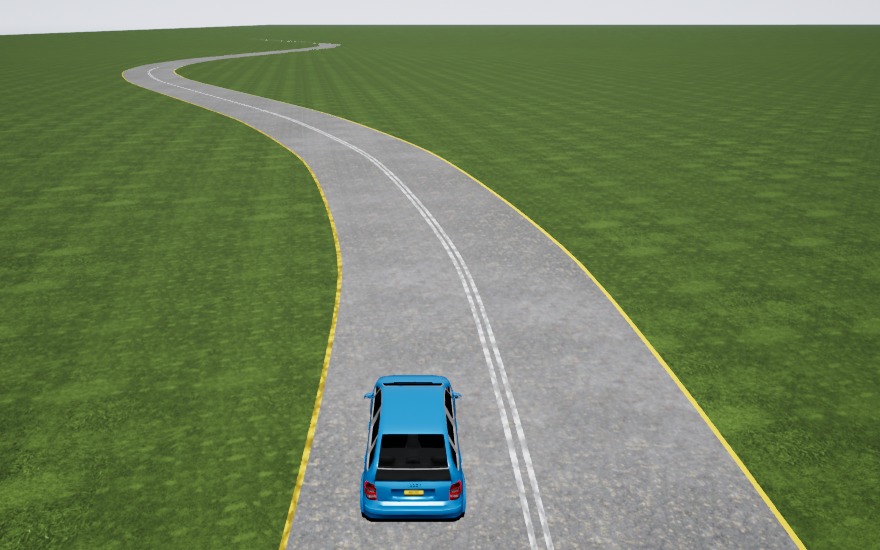}
     \end{subfigure}
     \quad
     \begin{subfigure}[b]{0.4\linewidth}
         \centering
         \includegraphics[width=\linewidth,trim={0 0 0 50px},clip]{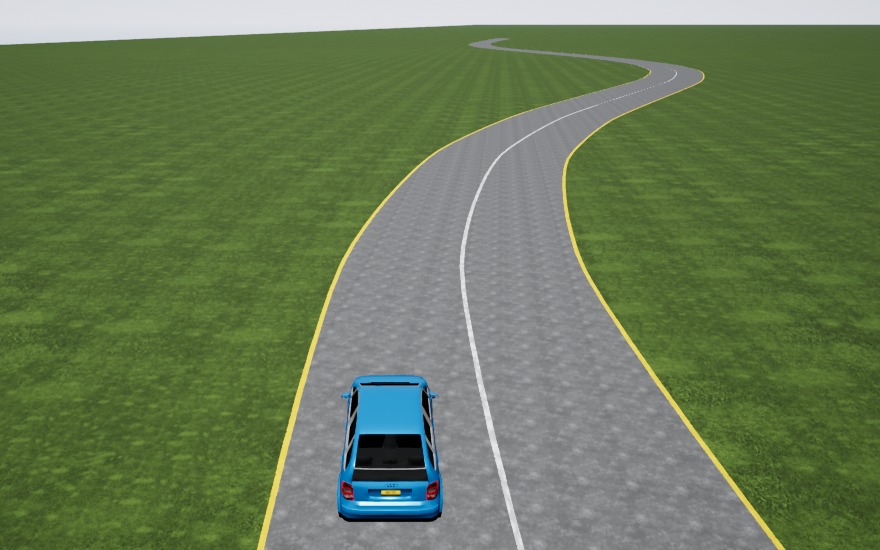}
     \end{subfigure}
\vspace{5mm}

     \begin{subfigure}[b]{0.4\linewidth}
         \centering
         \includegraphics[width=\linewidth,trim={0 0 0 50px},clip]{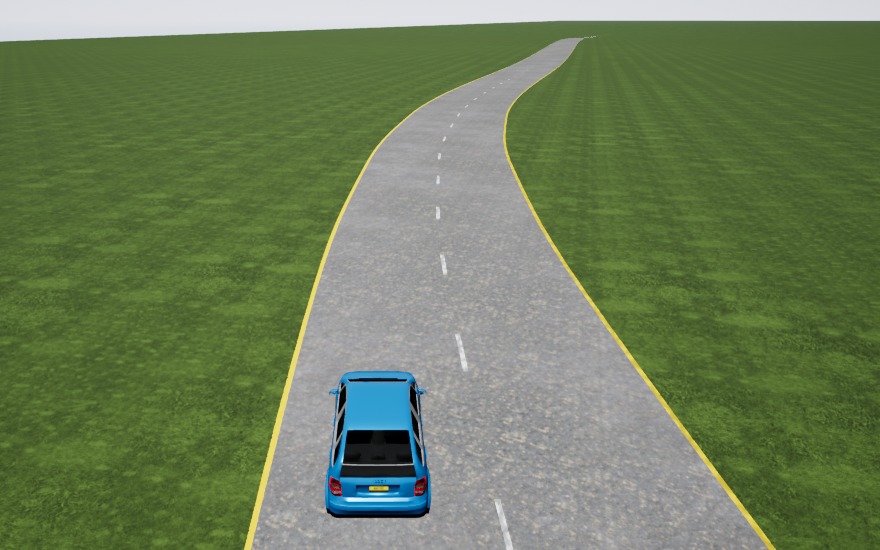}
     \end{subfigure}
     \quad
     \begin{subfigure}[b]{0.4\linewidth}
         \centering
         \includegraphics[width=\linewidth,trim={0 0 0 50px},clip]{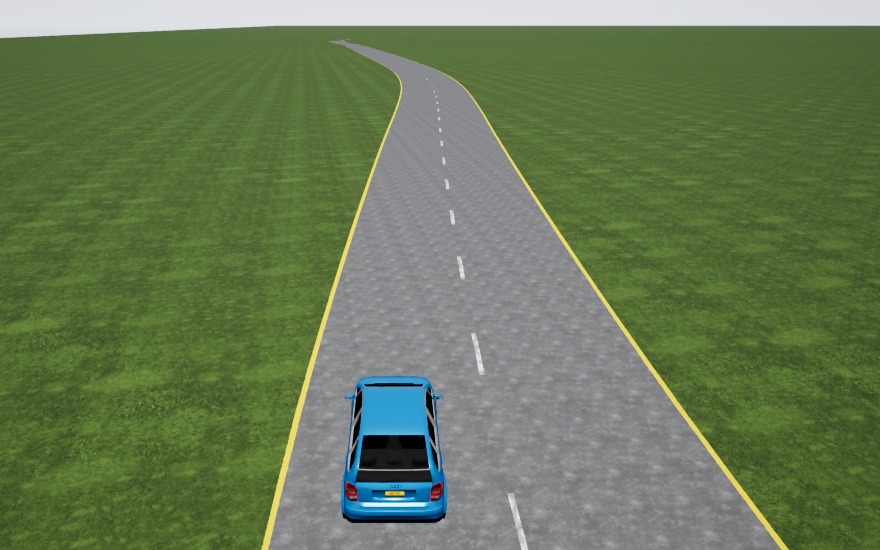}
     \end{subfigure}
\caption{Examples of different road environments randomly generated for each episode in our lane following simulator. We use procedural generation to randomly vary road texture, lane markings and road topology each episode. We train using a forward facing driver-view image as input.}
\label{fig:simulator}
\end{figure}

To test reinforcement learning algorithms in the context of lane following from image inputs we developed a 3D driving simulator, using Unreal Engine 4. It contains a generative model for country roads, supports varied weather conditions and road textures, and will in the future support more complex environments (see Figure \ref{fig:simulator} for game screenshots).

The simulator proved essential for tuning reinforcement learning parameters including: learning rates, number of gradient steps to take following each training episode and the correct termination procedure -- conservative termination leads to a better policy. It confirmed a continuous action space is preferable -- discrete led to a jerky policy -- and that DDPG is a suitable reinforcement learning algorithm. As described in the environment setup in Section~\ref{section:mdp}, reward granted in the simulator corresponded to the distance travelled before exiting lane, with new episodes resetting the car to the centre of the lane.

We found that we could reliably learn to learn follow in simulation from raw images within 10 training episodes. Furthermore, we found little advantage to using a compressed state representation (provided by a Variational Autoencoder). We found the following hyperparameters to be most effective, which we use for our real world experiments: future discount factor of 0.9, noise half-life of 250 episodes, noise parameters of $\theta$ of 0.6 and $\sigma$ of 0.4, 250 optimisation steps between episodes with batch size 64 and gradient clipping of 0.005.

\subsection{Real-world driving}
\label{section:real_world}
Our real world driving experiments mimic in many ways those conducted in simulation. However, executing this experiment in the real world is significantly more challenging. Many environmental factors cannot be controlled, and real-time safety and control systems must be implemented. For these experiments, we use a 250 meter section of road. The car begins at the start of the road to commence training episodes. When the car deviates from the lane and enters an unrecoverable position, the safety driver takes control of the vehicle ending the episode. The vehicle is then returned to the center of the lane to begin the next episode. We use the same hyperparameters that we found to be effective in simulation, with the noise model adjusted to give vehicle behaviour similar to that in simulation under the dynamics of the vehicle itself.

We conduct our experiments using a modified Renault Twizy vehicle, which is a two seater electric vehicle, shown in Figure~\ref{fig:teaser}. The vehicle weighs 500kg, has a top speed of 80 km/h and has a range of 100km on a single battery charge. We use a single monocular forward-facing video camera mounted in the centre of the roof at the front of the vehicle.
We use retrofitted electric motors to actuate the brake and steering, and electronically emulate the throttle position to regulate torque to the wheels. All computation is done on-board using a single NVIDIA Drive PX2 computer. The vehicle's drive-by-wire automation automatically disengages if the safety driver intervenes, either by using vehicle controls (brake, throttle, or steering), toggling the automation mode, or pressing the emergency stop. An episode would terminate when either speed exceeded 10km/h, or drive-by-wire automation disengaged, indicating the safety driver has intervened. The safety driver would then reset the car to the centre of the road and continue with the next episode.

\begin{figure}
\centering
     \begin{subfigure}[b]{0.7\linewidth}
         \centering
       \kern-2em
         \includegraphics[height=4.2cm]{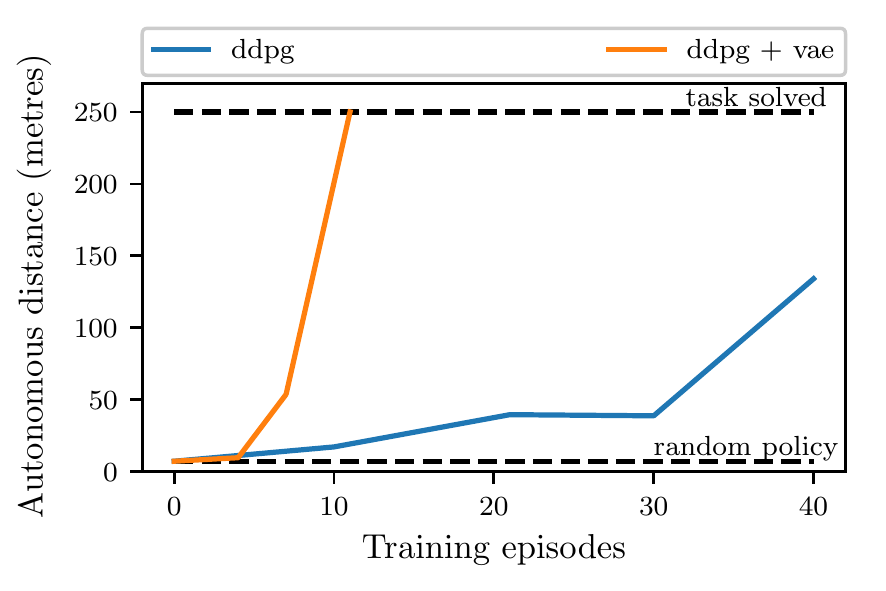}
         \caption[]{Algorithm results}
       \label{fig:algo-results}
     \end{subfigure}
     \begin{subfigure}[b]{0.28\linewidth}
         \centering
         \includegraphics[trim=0 0 200px 0, clip=true, height=3.2cm]{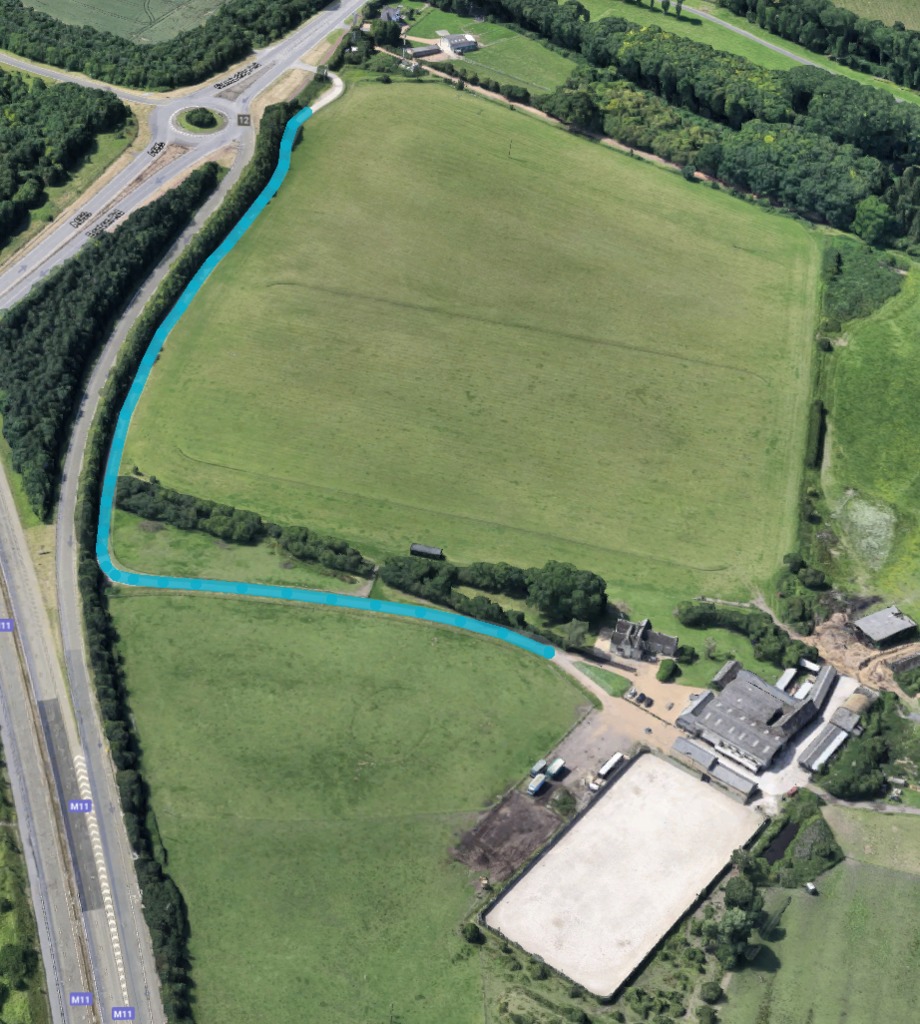}
         \vspace{7mm}
         \caption[]{Route}
        \label{fig:expt-rwd-route}
     \end{subfigure}
\caption{Using a VAE with DDPG greatly improves data efficiency in training over DDPG from raw pixels, suggesting that state representation is an important consideration for applying reinforcement learning on real systems. The 250m driving route used for our experiments is shown on the right.}
\label{fig:expt-rwd-route-vehicle}
\end{figure}


\begin{table*}[t]
\centering
 \begin{tabular}{l | c c c | c c}
 \hline
  & \multicolumn{3}{c|}{Training} & \multicolumn{2}{c}{Test}  \\ \hline
 Model & Episodes & Distance & Time & Meters per Disengagement & \# Disengagements \\
 \hline\hline
 Random Policy &-&-&-& 7.35 & 34 \\
 Zero Policy &-&-&-& 22.7 & 11 \\
 Deep RL from Pixels & 35 & 298.8 m & 37 min & 143.2 & 1 \\
 Deep RL from VAE & 11 & 195.5 m & 15 min & - & 0 \\
 \hline
 \end{tabular}
 \caption{Deep reinforcement learning results on an autonomous vehicle over a 250m length of road.  We report the best performance for each model. We observe the baseline RL agent can learn to lane follow from scratch, while the VAE variant is much more efficient, learning to succesfully drive the route after only 11 training episodes.}
\label{tbl:twizy-results}
\end{table*}




Table~\ref{tbl:twizy-results} shows the results of these experiments. Here, the major finding is that reinforcement learning can solve this problem in a handful of trials. Using 250 optimisation steps with batch size 64 took approximately 25 seconds, which made the experiment extremely manageable, considering manoeuvring the car to the centre of the lane to commence the next episode takes approximately 10 seconds anyway.
We also observe in the real world, where the visual complexity is much more difficult than simulation, a compressed state representation provided by a Variational Autoencoder trained online together with the policy greatly improved reliability of the algorithm. We compare our method to a zero policy (driving straight with constant speed) and random exploration noise, in order to confirm that the trial indeed required a non-trivial policy.
\footnote{A video of the training process for our vehicle learning to drive the 250m length of private road with the stateful RL training architecture (Section~\ref{section:stateful}) is available at \url{https://wayve.ai/blog/l2diad}}

\section{Discussion}

This work presents the first application of deep reinforcement learning to a full sized autonomous vehicle. The experiments demonstrate we are able to learn to lane follow with under thirty minutes of training -- all done on on-board computers.

In order to tune hyperparameters, we built a simple simulated driving environment where we experimented with reinforcement learning algorithms, maximising distance before a traffic infraction using DDPG as a canonical algorithm. The parameters found transferred amicably to the real-world, where we rapidly trained a policy to drive a real vehicle on a private road, with a reward signal consisting only of speed and termination upon control driver taking control. Notably, this reward requires no further information or maps of the environment. With more data, vehicles and larger models, this framework is general enough to scale to more complex driving tasks.

Whilst viable, this approach will require translation of reinforcement learning research advancements, as well as work on core reinforcement learning algorithms if it is to become a leading approach for scaling autonomous driving. We conclude by discussing our thoughts on the future work required.

In this work, we present a general reward function which asks the agent to maximise the distance travelled without intervention from a safety driver.  While this reward function is general, it has a number of limitations. It does not consider conditioning on a given navigation goal. Furthermore, it is incredibly sparse. As our agent improves, interventions will become significantly less frequent, resulting in weaker training signal. It is likely that further work is required to design a more effective reward function to learn a super-human driving agent. This will involve the careful consideration of many safety~\cite{amodei2016concrete} and ethical issues~\cite{thomson1985trolley}.

The second area for development suggested by the results here is a better state representation. Our experiments have shown that a simple Variational Autoencoder greatly improves the performance of DDPG in the context of driving a real vehicle. Beyond pixel-space autoencoders is a wealth of computer vision research addressing effective compression of images: here existing work in areas such as semantic segmentation, depth, egomotion and pixel-flow provide an excellent prior for what is important in driving scenes~\cite{kendall2017multi,badrinarayanan2017segnet,zhou2017unsupervised}. This research needs to be integrated with reinforcement learning approaches for real tasks, both model-free and model-based.

However, unsupervised state encoding alone will likely not be sufficient. In order to compress the state in a manner that makes it simple to learn a policy with just a small number of samples, information on which elements of the state (image observation) are important is required. This information should come from the reward and terminal signals. Reward and terminal information can be incorporated in an encoding in many ways, but one difficulty always prevails: credit allocation. Rewards obtained at a specific time step may be related to observations received many time-steps in the past. Thus good models used for this application will contain a temporal component.

Two areas that could greatly improve the availability of data for the application of reinforcement learning to real autonomous driving are semi-supervised learning~\cite{zhou2017unsupervised} and domain transfer~\cite{45924}. Whilst only a small portion of driving data might have rewards and terminals associated with it, as those are costly to obtain, the image embeddings -- and perhaps other aspects of models -- could benefit from driving data captured from dashcams in every-day vehicles. These could be used to pre-train the image autoencoder. In the context of a model-based RL system, these could also be used to approximate state transition functions, whilst advances in semi-supervised learning might allow us to utilise this data without reward/terminal labels data. Domain transfer, on the other hand, may allow us to create simulations sufficiently convincing that data from these may be used to train a policy that can be transferred directly onto a real car.

The algorithm used here is intentionally a common canonical approach, chosen to demonstrate the ease with which reinforcement learning may be applied to driving. Many improvements to it have been developed in the wider literature, including the use of natural gradients~\cite{schulman2015trust}. Other research has looked at better transformation of observations into states, typically using an RNN~\cite{hausknecht2015deep,igl2018deep}, as well as methods to perform multi-step planning, as in~\cite{farquhar2017tree}. It is no question that these could provide superior performance.

New advances in model-based reinforcement provide alternative exciting avenues for autonomous driving research, with work such as~\cite{deisenroth2011pilco} showing outstanding performance of models when observing directly the state of a physical system. This could offer significant benefits to an image-based domain. Alternative model-based approaches include~\cite{ha2018world} which learn to simulate episodes and learn in imagination.

We hope this paper inspires more research into applying reinforcement learning research to autonomous driving, perhaps combining it with elements from other machine learning techniques such as imitation learning and control theory. The method here solved a simple driving task in half an hour -- what more could be done in a day?


\clearpage
\bibliographystyle{IEEEtran}
\bibliography{references/rl,references/cv,references/robot}  

\begin{thebibliography}{10}
\providecommand{\url}[1]{#1}
\csname url@rmstyle\endcsname
\providecommand{\newblock}{\relax}
\providecommand{\bibinfo}[2]{#2}
\providecommand\BIBentrySTDinterwordspacing{\spaceskip=0pt\relax}
\providecommand\BIBentryALTinterwordstretchfactor{4}
\providecommand\BIBentryALTinterwordspacing{\spaceskip=\fontdimen2\font plus
\BIBentryALTinterwordstretchfactor\fontdimen3\font minus
  \fontdimen4\font\relax}
\providecommand\BIBforeignlanguage[2]{{%
\expandafter\ifx\csname l@#1\endcsname\relax
\typeout{** WARNING: IEEEtran.bst: No hyphenation pattern has been}%
\typeout{** loaded for the language `#1'. Using the pattern for}%
\typeout{** the default language instead.}%
\else
\language=\csname l@#1\endcsname
\fi
#2}}

\bibitem{badrinarayanan2017segnet}
V.~Badrinarayanan, A.~Kendall, and R.~Cipolla, ``Segnet: A deep convolutional
  encoder-decoder architecture for scene segmentation,'' \emph{IEEE
  Transactions on Pattern Analysis and Machine Intelligence}, 2017.

\bibitem{ort2018autonomous}
T.~Ort, L.~Paull, and D.~Rus, ``Autonomous vehicle navigation in rural
  environments without detailed prior maps,'' in \emph{International Conference
  on Robotics and Automation (ICRA)}, 2018.

\bibitem{sutton1998rli}
R.~S. Sutton and A.~G. Barto, \emph{Reinforcement Learning: An
  Introduction}.\hskip 1em plus 0.5em minus 0.4em\relax MIT Press, 1998.

\bibitem{silver2016go}
D.~Silver, A.~Huang, C.~J. Maddison, A.~Guez, L.~Sifre, G.~van~den Driessche,
  J.~Schrittwieser, I.~Antonoglou, V.~Panneershelvam, M.~Lanctot, S.~Dieleman,
  D.~Grewe, J.~Nham, N.~Kalchbrenner, I.~Sutskever, T.~P. Lillicrap, M.~Leach,
  K.~Kavukcuoglu, T.~Graepel, and D.~Hassabis, ``Mastering the game of go with
  deep neural networks and tree search,'' \emph{Nature}, vol. 529, no. 7587,
  pp. 484--489, 2016.

\bibitem{silver2017chess}
D.~Silver, T.~Hubert, J.~Schrittwieser, I.~Antonoglou, M.~Lai, A.~Guez,
  M.~Lanctot, L.~Sifre, D.~Kumaran, T.~Graepel, T.~P. Lillicrap, K.~Simonyan,
  and D.~Hassabis, ``Mastering chess and shogi by self-play with a general
  reinforcement learning algorithm,'' \emph{CoRR}, vol. abs/1712.01815, 2017.

\bibitem{mnih2015human}
V.~Mnih, K.~Kavukcuoglu, D.~Silver, A.~A. Rusu, J.~Veness, M.~G. Bellemare,
  A.~Graves, M.~Riedmiller, A.~K. Fidjeland, G.~Ostrovski, \emph{et~al.},
  ``Human-level control through deep reinforcement learning,'' \emph{Nature},
  vol. 518, no. 7540, p. 529, 2015.

\bibitem{gu2016robotic}
S.~Gu, E.~Holly, T.~Lillicrap, and S.~Levine, ``Deep reinforcement learning for
  robotic manipulation with asynchronous off-policy updates,'' in
  \emph{Robotics and Automation (ICRA), 2017 IEEE International Conference
  on}.\hskip 1em plus 0.5em minus 0.4em\relax IEEE, 2017, pp. 3389--3396.

\bibitem{lillicrap2015continuous}
T.~P. Lillicrap, J.~J. Hunt, A.~Pritzel, N.~Heess, T.~Erez, Y.~Tassa,
  D.~Silver, and D.~Wierstra, ``Continuous control with deep reinforcement
  learning,'' in \emph{International Conference on Learning Representations
  ({ICLR})}, 2016.

\bibitem{kanade1986autonomous}
T.~Kanade, C.~Thorpe, and W.~Whittaker, ``Autonomous land vehicle project at
  cmu,'' in \emph{Proceedings of the 1986 ACM fourteenth annual conference on
  Computer science}.\hskip 1em plus 0.5em minus 0.4em\relax ACM, 1986, pp.
  71--80.

\bibitem{wallace1985first}
R.~S. Wallace, A.~Stentz, C.~E. Thorpe, H.~P. Moravec, W.~Whittaker, and
  T.~Kanade, ``First results in robot road-following.'' in \emph{IJCAI}.\hskip
  1em plus 0.5em minus 0.4em\relax Citeseer, 1985, pp. 1089--1095.

\bibitem{montemerlo2008junior}
M.~Montemerlo, J.~Becker, S.~Bhat, H.~Dahlkamp, D.~Dolgov, S.~Ettinger,
  D.~Haehnel, T.~Hilden, G.~Hoffmann, B.~Huhnke, \emph{et~al.}, ``Junior: The
  stanford entry in the urban challenge,'' \emph{Journal of field Robotics},
  vol.~25, no.~9, pp. 569--597, 2008.

\bibitem{levinson2011towards}
J.~Levinson, J.~Askeland, J.~Becker, J.~Dolson, D.~Held, S.~Kammel, J.~Z.
  Kolter, D.~Langer, O.~Pink, V.~Pratt, \emph{et~al.}, ``Towards fully
  autonomous driving: Systems and algorithms,'' in \emph{Intelligent Vehicles
  Symposium (IV), 2011 IEEE}.\hskip 1em plus 0.5em minus 0.4em\relax IEEE,
  2011, pp. 163--168.

\bibitem{franke1998autonomous}
U.~Franke, D.~Gavrila, S.~Gorzig, F.~Lindner, F.~Puetzold, and C.~Wohler,
  ``Autonomous driving goes downtown,'' \emph{IEEE Intelligent Systems and
  Their Applications}, vol.~13, no.~6, pp. 40--48, 1998.

\bibitem{thrun2005probabilistic}
S.~Thrun, W.~Burgard, and D.~Fox, \emph{Probabilistic robotics}.\hskip 1em plus
  0.5em minus 0.4em\relax MIT press, 2005.

\bibitem{LinegarICRA2016}
C.~Linegar, W.~Churchill, and P.~Newman, ``Made to measure: Bespoke landmarks
  for 24-hour, all-weather localisation with a camera,'' in \emph{Proceedings
  of the IEEE International Conference on Robotics and Automation (ICRA)},
  Stockholm, Sweden, May 2016.

\bibitem{muller2006off}
U.~Muller, J.~Ben, E.~Cosatto, B.~Flepp, and Y.~L. Cun, ``Off-road obstacle
  avoidance through end-to-end learning,'' in \emph{Advances in neural
  information processing systems}, 2006, pp. 739--746.

\bibitem{geiger2012we}
A.~Geiger, P.~Lenz, and R.~Urtasun, ``Are we ready for autonomous driving? the
  kitti vision benchmark suite,'' in \emph{Proceedings of the IEEE Conference
  on Computer Vision and Pattern Recognition ({CVPR})}, 2012.

\bibitem{MaddernIJRR2016}
W.~Maddern, G.~Pascoe, C.~Linegar, and P.~Newman, ``1 year, 1000 km: The oxford
  robotcar dataset,'' \emph{The International Journal of Robotics Research},
  vol.~36, no.~1, pp. 3--15, 2017.

\bibitem{pomerleau1989alvinn}
D.~A. Pomerleau, ``Alvinn: An autonomous land vehicle in a neural network,'' in
  \emph{Advances in neural information processing systems}, 1989, pp. 305--313.

\bibitem{bojarski2016end}
M.~Bojarski, D.~Del~Testa, D.~Dworakowski, B.~Firner, B.~Flepp, P.~Goyal, L.~D.
  Jackel, M.~Monfort, U.~Muller, J.~Zhang, \emph{et~al.}, ``End to end learning
  for self-driving cars,'' \emph{arXiv preprint arXiv:1604.07316}, 2016.

\bibitem{sutton1998reinforcement}
R.~S. Sutton, A.~G. Barto, \emph{et~al.}, \emph{Reinforcement learning: An
  introduction}.\hskip 1em plus 0.5em minus 0.4em\relax MIT press, 1998.

\bibitem{deisenroth2011pilco}
M.~Deisenroth and C.~E. Rasmussen, ``Pilco: A model-based and data-efficient
  approach to policy search,'' in \emph{Proceedings of the 28th International
  Conference on machine learning (ICML)}, 2011, pp. 465--472.

\bibitem{williams2017information}
G.~Williams, N.~Wagener, B.~Goldfain, P.~Drews, J.~M. Rehg, B.~Boots, and E.~A.
  Theodorou, ``Information theoretic mpc for model-based reinforcement
  learning,'' in \emph{Robotics and Automation (ICRA), 2017 IEEE International
  Conference on}.\hskip 1em plus 0.5em minus 0.4em\relax IEEE, 2017, pp.
  1714--1721.

\bibitem{riedmiller2007learning}
M.~Riedmiller, M.~Montemerlo, and H.~Dahlkamp, ``Learning to drive a real car
  in 20 minutes,'' in \emph{Frontiers in the Convergence of Bioscience and
  Information Technologies, 2007. FBIT 2007}.\hskip 1em plus 0.5em minus
  0.4em\relax IEEE, 2007, pp. 645--650.

\bibitem{lecun1989backpropagation}
Y.~LeCun, B.~Boser, J.~S. Denker, D.~Henderson, R.~E. Howard, W.~Hubbard, and
  L.~D. Jackel, ``Backpropagation applied to handwritten zip code
  recognition,'' \emph{Neural computation}, vol.~1, no.~4, pp. 541--551, 1989.

\bibitem{kingma2013auto}
D.~P. Kingma and M.~Welling, ``Auto-encoding variational bayes,'' in \emph{The
  International Conference on Learning Representations ({ICLR})}, 2014.

\bibitem{rezende2014stochastic}
D.~J. Rezende, S.~Mohamed, and D.~Wierstra, ``Stochastic backpropagation and
  approximate inference in deep generative models,'' in \emph{Proceedings of
  the 31st International Conference on machine learning (ICML)}, 2014.

\bibitem{sutton2000policy}
R.~S. Sutton, D.~A. McAllester, S.~P. Singh, and Y.~Mansour, ``Policy gradient
  methods for reinforcement learning with function approximation,'' in
  \emph{Advances in neural information processing systems}, 2000, pp.
  1057--1063.

\bibitem{mnih2016asynchronous}
V.~Mnih, A.~P. Badia, M.~Mirza, A.~Graves, T.~P. Lillicrap, T.~Harley,
  D.~Silver, and K.~Kavukcuoglu, ``Asynchronous methods for deep reinforcement
  learning,'' in \emph{International Conference on Learning Representations
  ({ICLR})}, 2016.

\bibitem{schaul2015prioritized}
T.~Schaul, J.~Quan, I.~Antonoglou, and D.~Silver, ``Prioritized experience
  replay,'' in \emph{International Conference on Learning Representations
  ({ICLR})}, 2015.

\bibitem{uhlenbeck1930brownian}
G.~E. Uhlenbeck and L.~S. Ornstein, ``On the theory of the brownian motion,''
  \emph{Phys. Rev.}, vol.~36, pp. 823--841, Sep 1930.

\bibitem{amodei2016concrete}
D.~Amodei, C.~Olah, J.~Steinhardt, P.~Christiano, J.~Schulman, and D.~Man{\'e},
  ``Concrete problems in ai safety,'' \emph{arXiv preprint arXiv:1606.06565},
  2016.

\bibitem{thomson1985trolley}
J.~J. Thomson, ``The trolley problem,'' \emph{The Yale Law Journal}, vol.~94,
  no.~6, pp. 1395--1415, 1985.

\bibitem{kendall2017multi}
A.~Kendall, Y.~Gal, and R.~Cipolla, ``Multi-task learning using uncertainty to
  weigh losses for scene geometry and semantics,'' in \emph{Proceedings of the
  IEEE Conference on Computer Vision and Pattern Recognition ({CVPR})}, 2018.

\bibitem{zhou2017unsupervised}
T.~Zhou, M.~Brown, N.~Snavely, and D.~G. Lowe, ``Unsupervised learning of depth
  and ego-motion from video,'' in \emph{Proceedings of the IEEE Conference on
  Computer Vision and Pattern Recognition ({CVPR})}, 2017.

\bibitem{45924}
K.~Bousmalis, N.~Silberman, D.~Dohan, D.~Erhan, and D.~Krishnan, ``Unsupervised
  pixel-level domain adaptation with generative adversarial networks,'' in
  \emph{The IEEE Conference on Computer Vision and Pattern Recognition
  ({CVPR})}, 2017.

\bibitem{schulman2015trust}
J.~Schulman, S.~Levine, P.~Abbeel, M.~Jordan, and P.~Moritz, ``Trust region
  policy optimization,'' in \emph{International Conference on Machine
  Learning}, 2015, pp. 1889--1897.

\bibitem{hausknecht2015deep}
M.~Hausknecht and P.~Stone, ``Deep recurrent q-learning for partially
  observable mdps,'' \emph{CoRR, abs/1507.06527}, 2015.

\bibitem{igl2018deep}
M.~Igl, L.~Zintgraf, T.~A. Le, F.~Wood, and S.~Whiteson, ``Deep variational
  reinforcement learning for pomdps,'' in \emph{Proceedings of the 28th
  International Conference on machine learning (ICML)}, 2018.

\bibitem{farquhar2017tree}
G.~Farquhar, T.~Rockt{\"{a}}schel, M.~Igl, and S.~Whiteson, ``Treeqn and
  atreec: Differentiable tree planning for deep reinforcement learning,'' in
  \emph{International Conference on Learning Representations ({ICLR})}, 2018.

\bibitem{ha2018world}
D.~Ha and J.~Schmidhuber, ``World models,'' \emph{CoRR}, vol. abs/1803.10122,
  2018.

\end{thebibliography}

\end{document}